\documentclass[10pt,twocolumn,letterpaper]{article}

\usepackage{iccv}
\usepackage{times}
\usepackage{epsfig}
\usepackage{graphicx}
\usepackage{amsmath}
\usepackage{amssymb}
\usepackage{booktabs}
\usepackage{color}
\usepackage{multirow}
\usepackage{subfigure}
\usepackage{caption}
\usepackage{subfloat}
\usepackage{mathtools}
\usepackage{siunitx}
\usepackage{algorithm}
\usepackage{algorithmic}
\usepackage{bbm}
\usepackage{amsfonts}
\usepackage{mathrsfs}
\usepackage{multirow}
\usepackage[pagebackref=true,breaklinks=true,letterpaper=true,colorlinks,bookmarks=false]{hyperref}

\iccvfinalcopy % *** Uncomment this line for the final submission

\def\iccvPaperID{406} % *** Enter the CVPR Paper ID here

\begin{document}

%%%%%%%%% TITLE
\title{Left-right Discrepancy for Adversarial Attack on Stereo Networks}

\author{Pengfei Wang{$^{1}$},~ Xiaofei Hui{$^{2}$},~ Beijia Lu{$^{3}$},~ Nimrod Lilith{$^{1}$},~ Jun Liu{$^{2}$},~ Sameer Alam{$^{1}$}\\\vspace{-8pt}{\small~}\\
{$^{1}$}Nanyang Technological University;~~~ {$^{2}$}Singapore University of Technology and Design;~~~ \\
{$^{3}$}City University of Hong Kong \\
% {\small{{$^{\dag}$}Contribute equally~~{$^{*}$}Corresponding to: \tt{}}}
}

\maketitle
\thispagestyle{empty}

\begin{abstract}
Stereo matching neural networks often involve a Siamese structure to extract intermediate features from left and right images. The similarity between these intermediate left-right features significantly impacts the accuracy of disparity estimation. In this paper, we introduce a novel adversarial attack approach that generates perturbation noise specifically designed to maximize the discrepancy between left and right image features. Extensive experiments demonstrate the superior capability of our method to induce larger prediction errors in stereo neural networks, \eg outperforming existing state-of-the-art attack methods by 219\% MAE on the KITTI dataset and 85\% MAE on the Scene Flow dataset. Additionally, we extend our approach to include a proxy network black-box attack method, eliminating the need for access to stereo neural network. This method leverages an arbitrary network from a different vision task as a proxy to generate adversarial noise, effectively causing the stereo network to produce erroneous predictions. Our findings highlight a notable sensitivity of stereo networks to discrepancies in shallow layer features, offering valuable insights that could guide future research in enhancing the robustness of stereo vision systems.
\end{abstract}

\section{Introduction}
\label{sec:intro}
Stereo matching \cite{hirschmuller2007stereo} is a critical problem in computer vision tasks, including robotics and autonomous vehicles \cite{geiger2013vision, Geiger2012CVPR, maddern2016real}. Given a pair of rectified left and right images, the displacement between the matched pixels in left and right images is to be determined. In recent years, the accuracy of stereo disparity estimation has increased significantly due to availability of large open datasets and application of deep neural networks to stereo matching problems. However, deep neural networks are vulnerable to small perturbations in the input \cite{moosavi2016deepfool, peck2017lower}. Recently, some researchers have proposed investigating the robustness of a stereo network by applying adversarial attack techniques \cite{wong2021stereopagnosia, berger2022stereoscopic}. Adversarial attack is a technique designed to disrupt the neural network model to elicit incorrect predictions by the addition of imperceptible noise on the input left and right images.
% The most popular methods for adversarial attack are the fast gradient sign method (FGSM) \cite{goodfellow2014explaining} and its variants, which consider the adversarial attack as an optimization problem to maximize the difference between the network prediction result with the ground-truth. 
Existing works on adversarial attack of stereo network \cite{wong2021stereopagnosia} simply apply FGSM \cite{goodfellow2014explaining} and its variants \cite{kurakin2016adversarial, madry2017towards, dong2018boosting} to attack a stereo network by maximizing the loss function of the stereo network ($\eg$ smooth $L_1$ loss). However, such methods have not considered the unique architecture of the stereo networks.

\begin{figure}
    \centering
    \subfigure[Left image]{\includegraphics[width=0.48\linewidth]{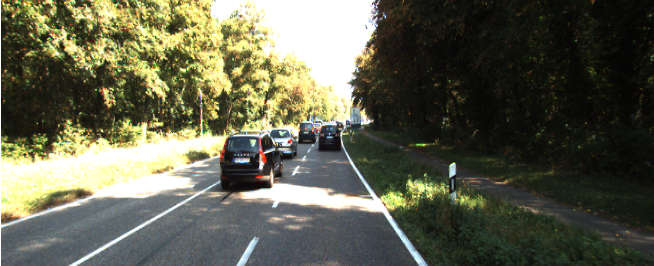}}
    \label{fig:1a}
    \hfill
    \subfigure[Right image]{\includegraphics[width=0.48\linewidth]{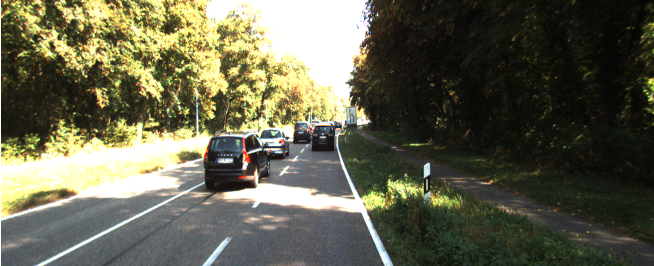}}
    \label{fig:1b}
    \vfill

    \subfigure[Disparity ground truth]{\includegraphics[width=0.48\linewidth]{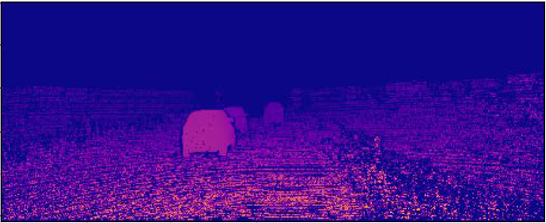}}
    \label{fig:1c}
    \hfill
    \subfigure[Disparity without attack]{\includegraphics[width=0.48\linewidth]{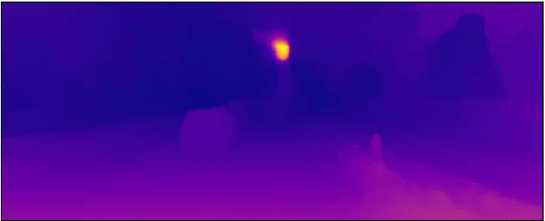}}
    \label{fig:1d}

    \vfill
    \subfigure[Predicted disparity after vanilla FGSM attack]{\includegraphics[width=0.48\linewidth]{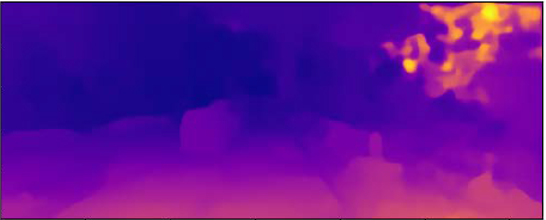}}
    \label{fig:1e}
    \hfill
    \subfigure[Predicted disparity after the proposed attack method]{\includegraphics[width=0.48\linewidth]{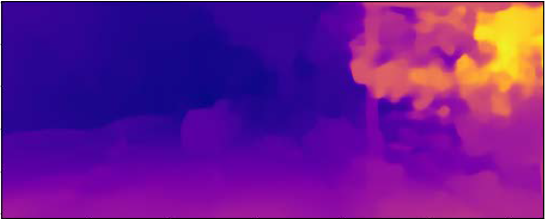}}
    \label{fig:1f}
  \caption{Illustration of the left-right discrepancy adversarial attack. The prediction results of the stereo network without adversarial attack show good performance. Compared to the existing attack method, the proposed attack method causes the stereo network to predict more significant error.}
  \label{fig:1}
\end{figure}
\vspace{-0.2cm}

Typically, the stereo network adopts a Siamese structure. For example, the DispNet \cite{mayer2016large} extracts high dimensional features for both left and right image with a weight sharing 2D CNN encoder, followed by a 2D CNN decoder to estimate the disparity in multi-resolution. AANet \cite{xu2020aanet} also extracts pyramid left and right features, followed by multi-scale cost aggregation for disparity regression.
% DeepPruner \cite{duggal2019deeppruner} uses spatial pyramid pooling (SPP) to extract left and right feature. The extracted features are fused by a customized PatchMatch \cite{bleyer2011patchmatch} operation to construct a cost volume, followed by 3D CNN to regress the disparity. 
Building upon such information, we make the assumption that the similarity between the left and right features have a significant effect on the performance of the estimated disparity. Following this, we propose a novel adversarial attack method for stereo networks. Specifically, we design a novel warping loss to generate adversarial noise by maximizing the dissimilarity between the intermediate left and right features of the stereo network. Intuitively, a larger left-right dissimilarity makes it more difficult for the network to regress correct disparity results. To evaluate our method, we have carried out extensive adversarial attack experiments to attack 3 popular disparity networks, AANet~\cite{xu2020aanet}, DeepPruner \cite{duggal2019deeppruner} and CREStereo \cite{li2022practical} on two datasets, Kitti 2015~\cite{geiger2015kitti}, and Scene Flow~\cite{mayer2016large}. The experiments demonstrate our method can significantly drive stereo networks to predict wrong disparity results (see Figure \ref{fig:1}), outperforming state-of-the-art adversarial attack methods noticeably.

As the target stereo network may not be always available, black-box attack is also crucial to investigate the robustness. Our method can also be easily extended to black-box attack for stereo networks. Existing black-box attack methods assume there is access to other stereo networks. Such methods apply adversarial attack on another available stereo network, such as AANet \cite{xu2020aanet}, generate noise, and apply on the target stereo network, such as DeepPruner \cite{duggal2019deeppruner}. Nevertheless, our proposed method does not require the availability of a stereo network. We can take any neural network from other vision tasks as the proxy network and generate adversarial noise to apply on stereo networks. 

In our experiments of both white-box attack and black-box attack, we have investigated the effect of leveraging different intermediate features in the proposed warping loss. We observe that using shallow features achieves better performance of attack. In addition, the analysis of left-right feature similarity statistics before and after adversarial attack indicates that using a shallow feature in the warping loss achieves the largest average similarity drop for the features before and after adversarial attack. Such results implies that stereo networks tend to be more sensitive to the discrepancy of shallow layer features.

Our contribution is summarized as follows:
\begin{itemize}
    \item We propose a novel loss function for adversarial attack on stereo network by maximizing the discrepancy on the left-right intermediate feature of stereo network;
    \item We perform exhaustive experiments to attack three popular stereo networks on two open datasets, and the performance outperforms state-of-the-art (SOTA) attack methods;
    \item We further extend our novel loss function to an efficient and practical proxy model black-box attack method without requiring access to any stereo network. Instead, our method can leverage an arbitrary network from other tasks and generate the adversarial noise to incur obvious prediction errors on stereo networks; And
    \item Our experiments reveal stereo networks are more sensitive to the discrepancy of shallow layer features, which can inspire further research on the robustness improvement of stereo networks.
\end{itemize}
\begin{figure*}[ht]
  \centering
%   \fbox{\rule{0pt}{2in} \rule{0.9\linewidth}{0pt}}
    \includegraphics[width=0.95\linewidth]{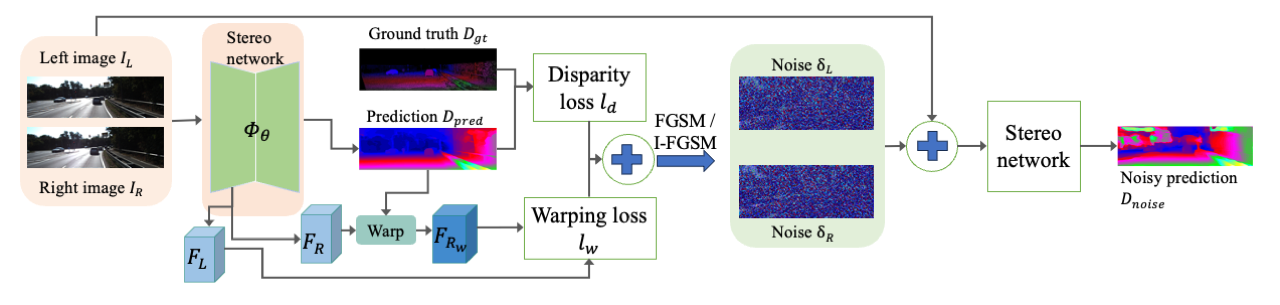}
   \caption{The architecture of the proposed attack method. The stereo network takes the left image $I_L$ and right image  $I_R$ as input, producing the estimated disparity $D_{est}$. We extract arbitrary intermediate left and right features $F_L, F_R$, and warp the $F_R$ based on the disparity (either ground truth Disparity $D_{gt}$ or the prediction $D_{pred}$ with clean images) to craft a pseudo feature $F_{R_w}$. A novel warping loss function $\ell_w$ is introduced to maximize the dissimilarity between the $F_L$ and $F_{R_w}$, which can be aggregated with the commonly used loss function of disparity $\ell_{d}$. The adversarial noise $\delta_L, \delta_R$ is generated by maximizing the loss function using the FGSM \cite{goodfellow2014explaining} or I-FGSM \cite{kurakin2016adversarial} algorithm. The generated noise $\delta_L, \delta_R$ is added to the input $I_L, I_R$, resulting in the noisy predicted disparity result $D_{noise}$.}
   \label{fig:architecture}
\end{figure*}
\vspace{-0.2cm}
\section{Related Works}
\label{sec:related}

\subsection{White-box Attack on Stereo Network}
Adversarial attack \cite{szegedy2013intriguing} has been widely investigated in classification tasks. By adding small imperceptible noise on the input image, the prediction results of neural networks can alter significantly. FGSM \cite{goodfellow2014explaining} is a popular attack algorithm that adds noise along the signed gradient to maximize the loss function. Some researchers have proposed boosting the attack performance by extending FGSM \cite{goodfellow2014explaining} in an iterative manner \cite{kurakin2016adversarial}. Other than classification task, adversarial attack was extended to dense-pixel tasks, including segmentation \cite{arnab2018robustness, xie2017adversarial, nesti2022evaluating}, optical flow \cite{ranjan2019attacking, schmalfuss2022perturbation, schrodi2022towards}, and monocular depth estimation \cite{cheng2022physical, zhang2020adversarial, wong2020targeted}. Recently, adversarial attacks on stereo networks has also been investigated \cite{wong2021stereopagnosia, berger2022stereoscopic} by applying FGSM and its variants to the stereo network by maximizing the error between the estimated disparity with the ground truth. However, they have not considered the characteristics of stereo networks. The stereo network relies on the matching of pixels between the left and right images. The similarity between the left and right features are crucial to the prediction performance of a stereo network. Our proposed methods generate adversarial noise by introducing a novel loss to increase the left-right feature discrepancy. Through this strategy, the stereo network is fooled to make much more significant error.

\subsection{Black-box Attack on Stereo Network}
The target neural network is not always available for adversarial attack. In such cases, black-box attack can be used to optimize the adversarial noise to mislead the output of the neural network. A popular black-box strategy was proposed based on the transferability of adversarial examples \cite{cheng2019improving, naseer2019cross}. Such a strategy was also adapted to black-box attack attack on stereo networks \cite{wong2021stereopagnosia,berger2022stereoscopic}. The adversarial noise generated by attacking one stereo network, such as AANet \cite{xu2020aanet} was applied on another stereo network, such as DeepPruner \cite{duggal2019deeppruner} to investigate the transferability of adversarial noise among different stereo networks. However, such a strategy has the limitation of assuming access to at least one stereo network to generate adversarial noise. In contrast, our method eliminates this requirement by generating adversarial noise with any neural network in other tasks, such as the basic ResNet-18 \cite{he2016deep} of image classification. Our approach establishes a crucial knowledge connection between the stereo network and other tasks, simplifies the generation of adversarial noise and enhances the transferability.

\subsection{Stereo Disparity Estimation}
% Stereo disparity estimation computes the disparity of each pixel between a pair of rectified stereo images. Finding local correspondences between two images plays a key role in the generation of high-quality disparity maps. 
Deep learning was proposed to be used in stereo matching \cite{zbontar2015computing} by training a Siamese network to extract the features from patches and subsequently perform conventional patch comparison based on the deep features. DispNet \cite{mayer2016large} introduced an end-to-end CNN with a Siamese structure to extract deep features from left and right images, and constructed a cost volume with the extracted features to predict the disparity. Following a similar approach, AANet \cite{xu2020aanet} also employed a Siamese structure to extract features from left and right images, utilizing deformable convolution for adaptive sampling in cost aggregation. DeepPruner \cite{duggal2019deeppruner} designed a differentiable patch matching to construct the cost volume from deep features extracted by a Siamese structure. CREStereo \cite{li2022practical} designed a hierarchical network with RNN to update disparity from coarse to fine.

Observing the significant influence of left-right feature similarity, certain approaches proposed incorporating a sub-network to fine-tune the disparity results through augmenting left-right similarity. LRCR~\cite{jie2018left} proposed comparing the left features and warped right features to generate an error map, and feed the error map with the cost volume to a convolution LSTM model to refine the disparity iteratively. PCW-Net~\cite{shen2022pcw} similarly adopts the concept of disparity refinement by concatenating the left feature, the warped right feature and the difference between them to construct the cost volume for refinement. In summary, the similarity between the warped right image feature and the left image feature emerges as a crucial cue for enhancing the disparity accuracy through locating the estimated error of the disparity. This observation aligns closely with our assumption that an increase in left-right discrepancy directly contributes to larger prediction errors in disparity prediction.
\section{Methods}
\label{sec:methods}
Most stereo matching networks take the Siamese structure to extract pyramid intermediate features from both left and right images with a weight-sharing encoder. Then, the left and right features are fused to build a bottleneck feature, followed by a 2D CNN or 3D CNN decoder to estimate the disparity results. During training, the loss function back propagates the gradient through the pyramid left and right features to the input image. This kind of stereo matching structure leverages the similarity between the left and right features to predict accurate disparity results. Intuitively, assume the disparity of the pixel $I_L(u, v)$ in left image is $d$, so this pixel corresponds to the pixel $I_R(u-d, v)$ in the right image. The image patch around the pixel $I_L(u, v)$ and $I_R(u-d, v)$ show a similar appearance. As a convolution network keeps local connectivity, the extracted features of $I_L(u, v)$ and $I_R(u-d, v)$ also tend to be similar. Inspired by this observation, we propose an assumption that the prediction accuracy of the stereo network is highly related to the similarity between the left and right feature extracted by the neural network. The performance of the network will become worse if this similarity decreases. Based on this assumption, we propose a novel adversarial attack method on stereo networks by increasing the dissimilarity between the left and right intermediate features. 

\subsection{White-box Attack}
Given a pair of left and right stereo images $I_L, I_R \in R^{W \times H \times 3}$) and a pre-trained stereo model $\Phi_{\theta}$, the adversarial attack adds imperceptible noise $\delta_L, \delta_R (||\delta_L||_{\infty} < \epsilon, ||\delta_R||_{\infty} < \epsilon)$ to both the two images, to generate distinct error on the predicted disparity by maximizing the loss function $\ell(\Phi_{\theta}(I_L + \delta_L, I_R + \delta_R), D)$, where $D$ is the disparity (either ground truth or prediction with clean images).
% \begin{equation}
%     \begin{aligned}
%         \max_{\delta_L, \delta_R} \quad & \ell(\Phi_{\theta}(I_L + \delta_L, I_R + \delta_R), D) \\
%         % s.t. \quad & ||\delta_L||_{\infty} < \epsilon \\ 
%         %            & ||\delta_R||_{\infty} < \epsilon
%     \end{aligned}
% \end{equation}

% where $D$ is the disparity (either ground truth or prediction with clean images).
% Through maximizing this loss function, a pair of perturbation noise is crafted.

% \subsection{Algorithm}
% As shown in Fig. \ref{fig:architecture}, a general stereo network takes the left image $I_L$ and right image $I_R$ as input, and extract intermediate features $F_i_L, F_i_R, i=1,2...$ in multiple scales by an encoder $\Phi_{\theta, E}$. The feature size of scale $i$ is $B \times C_i \times H_i \times W_i$. Then, a decoder $\Phi_{\theta, D}$ takes the intermediate features $F_i_L, F_i_R$ to predict the disparity $D_{pre}$. The performance of the stereo network is evaluated by comparing the error between $D^{pre}$ the predictied disparity of the pretrained model and the ground truth disparity $D_{gt}$. The ground truth disparity is resized to $D_{gt, i}$ at each scale $i$ corresponding the the intermediate features. Based on the resized disparity, the right features ${F_i_R, i=1,2..}$ is warped to generate a pseudo feature ${F_i_{R_w}, i=1,2..}$ as Equation \ref{eq:warp_feature}.
A typical stereo network adopts an encoder-decoder architecture. The encoder extracts intermediate features in multiple scales from the left and right images. Subsequently, the decoder utilizes the intermediate features to estimate the probability of the disparity for each pixel. As the disparity indicates the correspondence between the pixels in the left and right images, predicting the disparity can be conceptualized as a search for the optimal matching pixel pair. The effectiveness of search is notably affected by the similarity between the left and right intermediate features, further impacting the robustness of the stereo network. Previous methods \cite{wong2021stereopagnosia, berger2022stereoscopic} investigate the robustness of stereo network by incorporating FGSM \cite{goodfellow2014explaining} adversarial attack or its variants to increase the disparity loss function $\ell_d$ of the stereo network, such as smooth $L_1$ loss between the predicted disparity and the ground truth. Such methods neglect the significance of the left-right feature similarity in the stereo network. Our method introduces an innovative loss function to explicitly examine the impact of the left-right feature similarity on the robustness of stereo networks (see Figure \ref{fig:architecture}). In particular, we can arbitrarily extract one layer of left and right intermediate features $F_L, F_R \in R^{W_F \times H_F \times C_F}$ from the stereo network $\Phi_{\theta}$. Subsequently, the right feature is warped based on the disparity $D$ (either ground truth or prediction with clean images) to align with left feature. To account for the resolution mismatch between the feature and the disparity, the disparity is scaled to $D_{F}$ at the same resolution with the right feature. Based on the resized disparity $D_{F}$, the right feature ${F_R}$ is warped to generate a pseudo feature ${F_{R_w}}$.

Intuitively, if the stereo network performs well, the left feature and the pseudo feature should be quite similar. The similarity between the left feature $F_L$ and the pseudo feature $F_{R_w}$ is defined as the dot product between the two features. We use the dissimilarity of left and right feature as the loss function $\ell_{w}$ (as Equation \ref{eq:obj_warp}) to optimize the adversarial noise for the input left and right images. This loss function can be used in conjunction with the disparity loss $\ell_{d}$.

\vspace{-0.4cm}
\begin{equation}
    \begin{aligned}
        & D_{F}(u, v) = \frac{W_F}{W}D(\frac{ u \cdot W_F }{ W }, \frac{ v \cdot H_F }{ H }) \\
        & F_{R_w}(u, v) = F_R(u + D(u, v), v) \\
        & \ell_w(F_L, F_{R_w}, D) = 1 - \frac{F_L}{||F_L||} \cdot \frac{F_{R_w}}{||F_{R_w}||} \\
        & \ell(I_{L}, I_{R}, D) = \ell_{d}(I_{L}, I_{R}, D) + \lambda \ell(F_L, F_{R_w}, D)
    \end{aligned}
    \label{eq:obj_warp}
\end{equation}

\subsection{Proxy Network Black-box Attack}
Black-box attack can be adopted when the target stereo network model is unavailable. A general method of black-box attack is to apply an adversarial attack on one accessible stereo network model (for example, CREStereo \cite{li2022practical}), generating the adversarial noise off-the-self, then apply on another stereo network without access (for example, DeepPruner \cite{duggal2019deeppruner}, AANet \cite{xu2020aanet}). However, there is a requirement to have access to at least one stereo network model. We propose a novel proxy network black-box attack method which does not require access to a stereo network model. In our method, we can take a neural network from another task, for example, the ResNet-18 \cite{he2016deep} pretrained on a basic image classification task. Specifically, we take an arbitrary pretrained network model $\Phi'( \theta')$ as a proxy network. The proxy intermediate features $F_{L'}, F_{R'}$ can be obtained by feeding the left and right images $I_{L}, I_{R}$ into the the proxy network successively. The adversarial noise can be generated by maximizing the warping loss as Equation \ref{eq:proxy_obj}. 

\vspace{-0.6cm}
\begin{equation}
\begin{aligned}
    & \ell'(I_{L}, I_{R}) = \ell_w({F'_L}, F'_{R_w}, D)= 1 - \frac{F'_L}{||F'_L||} \cdot \frac{F'_{R_w}}{||F'_{R_w}||} \\
    & D'_{F}(u, v) = \frac{W_F}{W} \cdot D(\frac{ u \cdot W_F }{ W }, \frac{ v \cdot H_F }{ H }) \\
    & F'_{R_w}(u, v) = F'_R(u + D'_F(u, v), v)
    % F_{L'} & = \Phi'_{\theta'}(I_{L}) \\
    % F_{R'} & = \Phi'_{\theta'}(I_{R})
    \label{eq:proxy_obj}
\end{aligned}
\end{equation}
\vspace{-0.6cm}

\section{Experiments}
\label{sec:exp}

To evaluate the performance of the attack methods, Mean Absolute Error (MAE), Root Square Mean Error and D1-error are adopted for measurement. The calculations of the metrics are in the Supplementary. The proposed adversarial attack method is evaluated on two datasets, KITTI 2015 \cite{geiger2015kitti} and Scene Flow dataset \cite{mayer2016large}. Three stereo networks are used as the target network for adversarial attack, AANet \cite{xu2020aanet}, DeepPruner \cite{duggal2019deeppruner}, and CREStereo \cite{li2022practical}.

% \textbf{KITTI 2015} is an open dataset with 200 pair of training stereo images with groundtruth, and 200 testing pairs without groundtruth. The images are at a resolution of $1242 \times 375$, collected with a vehicle in outdoor environment. Following \cite{wong2021stereopagnosia}, we have tested the attack results on all the 200 pair of training images in the experiments.

% \textbf{Scene Flow dataset} is a large scale synthetic dataset with 34,803 training and 4,248 testing images at a resolution of $960 \times 540$. Most of the stereo neural networks are trained on the scene flow training set and evaluating on the testing set. In our experiments, we have evaluated the performance of adversarial attack on the testing set with 4,248 pair of stereo images. 

\subsection{Comparison with State-of-the-art Methods for White-box Attack}
\label{subsec:compare}

\begin{figure}[htbp]
    \centering
   \subfigure[Left image]{\includegraphics[width=0.48\linewidth]{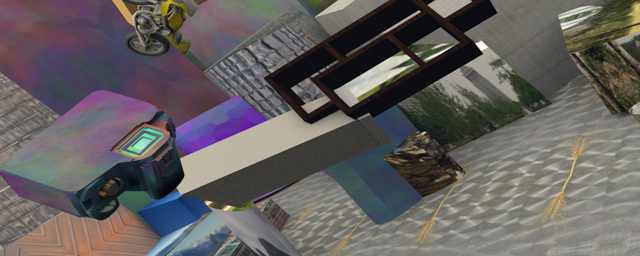}}
    \label{fig:white_aanet_sf_left_fgsm}
  \hfill
   \subfigure[Clean Estimation]{\includegraphics[width=0.48\linewidth]{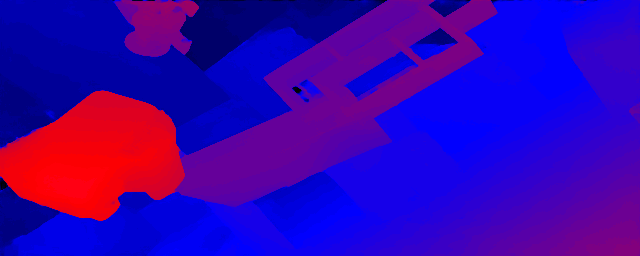}}
    \label{fig:white_aanet_sf_clean}
  \hfill
  
   \subfigure[Vanilla One-shot]{\includegraphics[width=0.48\linewidth]{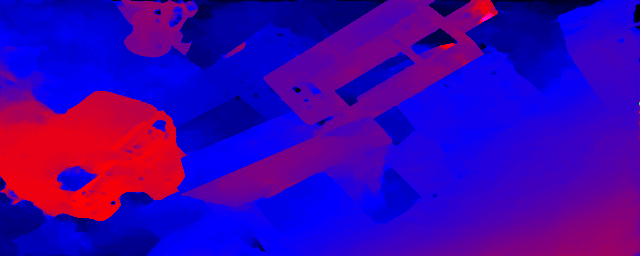}}
    \label{fig:white_aanet_sf_vanilla_fgsm}
  \hfill
   \subfigure[Vanilla Iterative]{\includegraphics[width=0.48\linewidth]{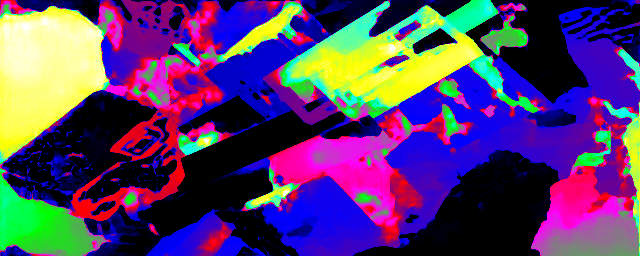}}
    \label{fig:white_aanet_sf_vanilla_ifgsm}
  \hfill
  
   \subfigure[Our One-shot]{\includegraphics[width=0.48\linewidth]{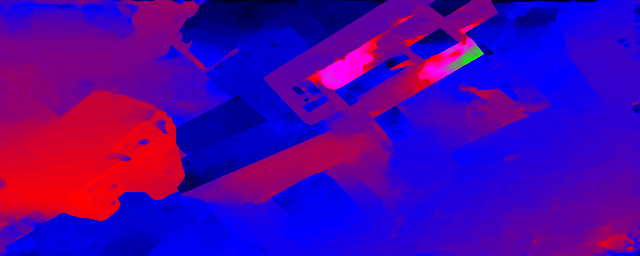}}
    \label{fig:white_aanet_sf_ours_fgsm}
  \hfill
   \subfigure[Our Iterative]{\includegraphics[width=0.48\linewidth]{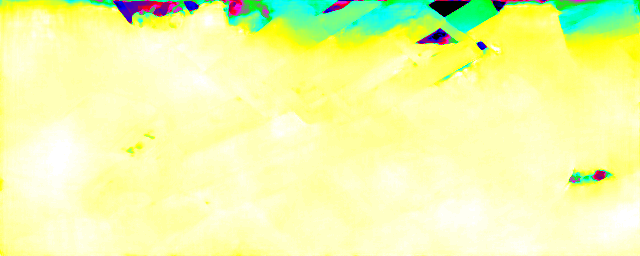}}
    \label{fig:white_aanet_sf_vanilla_ifgsm}
    \caption{White-box attack results on AANet with Scene Flow dataset}
    \label{fig:color_aanet_sf_white}
\end{figure}

Table \ref{tab:compare} compares our attack method with the state-of-the-art by attacking the three stereo networks on the two datasets. The SOTA method \cite{wong2021stereopagnosia} only uses the disparity loss $\ell_d$ (\eg the smooth $L_1$ loss) to generate adversarial noise with one-shot attack algorithm (FGSM) and iterative attack algorithm (I-FGSM). Our method uses a combination of disparity loss $\ell_d$ and the proposed novel warping loss $\ell_{w}$. To make the comparison fair, we have compared the results of the one-shot and iterative algorithms separately. As the results in Table \ref{tab:compare} show, our proposed method achieves better performance than the SOTA. Specifically, in the adversarial attack experiments on the target stereo network AANet and CREStereo, our method can drive both the stereo networks towards much more significant errors in terms of all three evaluation metrics. Comparing the errors caused by the adversarial attack on the three evaluation metrics, CREStereo \cite{li2022practical} is the most robust against the perturbation, while DeepPruner \cite{duggal2019deeppruner} is the least. The relatively higher robustness of CREStereo might be due to the fact that this model applies a hierarchical structure to refine the disparity from coarse to fine. During the forward propagation of the network, the effect of the perturbation signal is reduced by the refinement operations. The prediction error caused by the one-shot algorithm on CREStereo is limited. While using iterative algorithm, our joint loss $\ell$ induces substantially greater error than the vanilla disparity loss $\ell_w$ across all the three metrics. Specifically, we observe a 219\% increase in MAE, a 123\% increase in RMSE and a 13\% increase in D1-error on KITTI dataset. Similarly, on Scene Flow dataset, our joint loss achieves an 85\% higher MAE, a 59\% higher RMSE, and a 13\% higher D1-error. For the AANet \cite{xu2020aanet}, our joint loss $\ell$ can incur much greater error than the vanilla loss $\ell_d$ using both one-shot and iterative algorithms. For the DeepPruner \cite{duggal2019deeppruner}, the performance of adversarial attack is on par with the vanilla loss. This might be because the DeepPruner model is quite vulnerable, both our loss and the vanilla loss can easily cause significant errors on the predicted disparity results. Some results are visualized in Figure \ref{fig:color_aanet_sf_white}. The vanilla one-shot attack causes the AANet to make small error prediction on the left flying object (Figure \ref{fig:color_aanet_sf_white}e), while our one-shot attack causes more obvious errors on larger regions. The vanilla iterative attack drives the network to make significant errors on the entire image, while the structure of some objects is still clear. However, our iterative attack leads the network to estimate entirely incorrect results and the scene structure is no longer discernible. More visualization results are found in the Supplementary.
% As in Figure \ref{fig:color_kitti_cre_white}, the CREStereo is quite robust to one-shot attack. While applying iterative attack, 

\begin{table*}[htbp]
    \centering
    \caption{Comparison between our method with state-of-the-art by attacking three stereo networks (AANet \cite{xu2020aanet} and DeepPruner \cite{duggal2019deeppruner} and CREStereo \cite{li2022practical}) on KITTI 2015 dataset and Scene Flow dataset. Vanilla is the method using only the the disparity loss $\ell_d$ of the stereo network. Our method uses a combination of warping loss $\ell_w$ and disparity loss $\ell_d$.}
\begin{tabular}{cclllllll}
\multirow{2}{*}{\textbf{Target}}      & \multicolumn{2}{c}{\multirow{2}{*}{\textbf{Method}}}   & \multicolumn{3}{c}{\textbf{KITTI 2015}}               & \multicolumn{3}{c}{\textbf{Scene Flow}}               \\
                                      & \multicolumn{2}{c}{}                                   & \textbf{MAE}    & \textbf{RMSE}   & \textbf{D1-error} & \textbf{MAE}    & \textbf{RMSE}   & \textbf{D1-error} \\
\hline
\multirow{5}{*}{\textbf{AANet}}       & \multicolumn{2}{c}{None}                      & 0.42            & 0.99            & 1.11              & 1.81            & 4.91            & 5.25              \\
                                      & \multirow{2}{*}{One-shot}   & Vanilla & 2.91            & 4.01            & 34.36             & 5.88            & 9.95            & 46.90             \\
                                      &                                     & Ours    & \textbf{21.51}  & \textbf{32.12}  & \textbf{65.41}    & \textbf{9.55}   & \textbf{14.70}  & \textbf{53.03}    \\
                                      & \multirow{2}{*}{Iterative} & Vanilla & 48.36           & 62.35           & 82.77             & 51.52           & 70.15           & 88.78             \\
                                      &                                     & Ours    & \textbf{163.69} & \textbf{165.89} & \textbf{99.47}    & \textbf{138.39} & \textbf{146.21} & \textbf{98.24}    \\
\hline
\multirow{5}{*}{\textbf{DeppPrunner}} & \multicolumn{2}{c}{None}                      & 0.45            & 0.98            & 1.13              & 4.65            & 10.76           & 19.69             \\
                                      & \multirow{2}{*}{One-shot}   & Vanilla & 4.59            & 5.99            & \textbf{56.76}    & 13.88           & 24.48           & 74.17             \\
                                      &                                     & Ours    & \textbf{5.89}   & \textbf{8.92}   & 51.62             & \textbf{14.23}  & \textbf{25.14}  & \textbf{74.21}    \\
                                      & \multirow{2}{*}{Iterative} & Vanilla & 179.13          & 179.38          & \textbf{100.00}   & 170.71          & 172.08          & \textbf{99.99}    \\
                                      &                                     & Ours    & \textbf{179.34} & \textbf{179.59} & \textbf{100.00}   & \textbf{170.73} & \textbf{172.11} & \textbf{99.99}    \\
\hline
\multirow{5}{*}{\textbf{CREStereo}}   & \multicolumn{2}{c}{None}                      & 0.61            & 1.51            & 2.92              & 0.81            & 3.03            & 3.12              \\
                                      & \multirow{2}{*}{One-shot}   & Vanilla & 1.66            & 2.55            & 7.80              & 2.10            & 4.52            & 7.74              \\
                                      &                                     & Ours    & \textbf{1.70}   & \textbf{2.70}   & \textbf{14.33}    & \textbf{2.30}   & \textbf{4.86}   & \textbf{17.47}    \\
                                      & \multirow{2}{*}{Iterative} & Vanilla & 26.43           & 40.88           & 82.77             & 46.65           & 65.44           & 82.36             \\
                                      &                                     & Ours    & \textbf{84.36}  & \textbf{91.35}  & \textbf{93.68}    & \textbf{86.16}  & \textbf{104.09} & \textbf{93.26}   
\end{tabular}
    \label{tab:compare}
\end{table*}

% \usepackage{multirow}

\subsection{Effect of White-box Attack on Scene Geometry}
\begin{figure}[htbp]
\centering
   \subfigure[One-shot]{\includegraphics[width=0.8\linewidth]{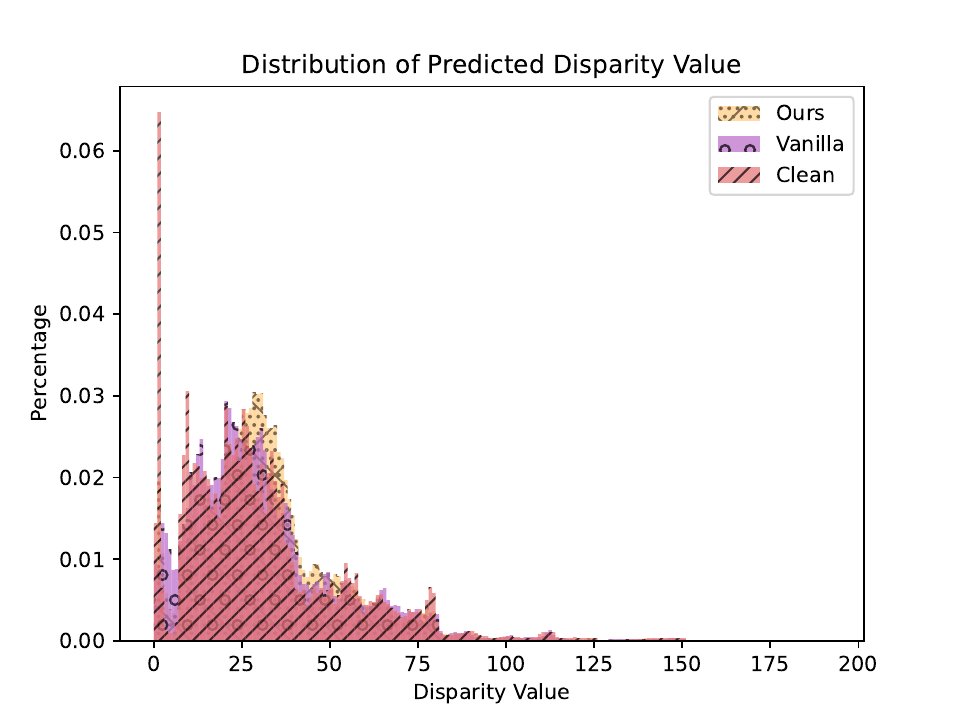}}
    \label{fig:hist_one-shot}
    
   \subfigure[Iterative]{\includegraphics[width=0.8\linewidth]{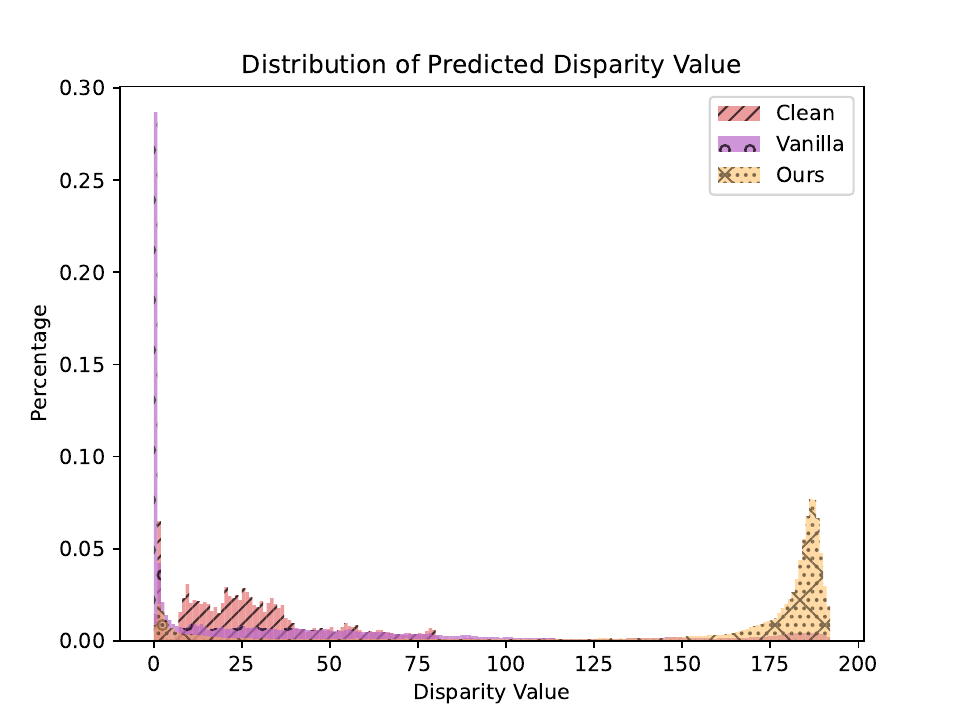}}
    \label{fig:hist_iterative}

    \caption{Distribution of disparity before and after adversarial attack. ``Clean" represents the disparity prediction of the stereo network without adversarial attack, ``Vanilla" and ``Ours" represents the disparity prediction after adversarial attack by vanilla loss $\ell_d$ and our joint loss function $\ell$, respectively.}
    \label{fig:hist}
\end{figure}

Based on the predicted disparity of the stereo network, it is straightforward to calculate the distance from the camera to any 3D points in the scene, which is useful in obstacle avoidance of autonomous driving, robot navigation, etc. Thus, the robustness of the stereo network to estimate the scene geometry is of great importance. To analyze how the proposed white-box attack method affects the estimated scene geometry, following \cite{berger2022stereoscopic}, we performed a statistical examination of the disparity values across the dataset. As Figure \ref{fig:hist}a shows, the vanilla one-shot attack moves the peak of distribution from 0 to $\approx$ 30px, while ours moves the peak to $\approx$ 40px. Such a effect is more significant when using iterative attack. As Figure \ref{fig:hist}b shows, the vanilla iterative attack shifts the peak of disparity about $\approx$ 30px, while ours moves the peak about $\approx$ 180px. Thus, our attack method makes the predicted scene to be closer to the stereo camera (larger disparity means smaller distance). 

% In addition, to analyze how the attack method affect the estimation results on objects at different distances, the estimation error on various range of disparities are summarized as in Figure \ref{fig:hist_error}. 

% \begin{figure}[htbp]
%     \centering
    
%    \begin{subfigure}{0.48\linewidth}
%     % \fbox{\rule{0pt}{2in} \rule{.9\linewidth}{0pt}}
%     \includegraphics[width=0.96\linewidth]{fig/hist/error_aanet_sf_fgsm.png}
%     \caption{One-shot}
%     \label{fig:hist_one-shot}
%   \end{subfigure}
%   \hfill
%    \begin{subfigure}{0.48\linewidth}
%     % \fbox{\rule{0pt}{2in} \rule{.9\linewidth}{0pt}}
%     \includegraphics[width=0.96\linewidth]{fig/hist/error_aanet_sf_ifgsm.png}
%     \caption{Iterative}
%     \label{fig:hist_iterative}
%   \end{subfigure}
  
%     \caption{Distribution of disparity error before and after adversarial attack}
%     \label{fig:hist_error}
% \end{figure}

\subsection{Ablation Study on the Warping Loss}
In our adversarial attack method, an arbitrary layer of intermediate feature is utilized in the warping loss $\ell_w$ to generate noise. To investigate how the attack performance is affected by taking different layers of intermediate features in the warping loss $\ell_w$, we have performed an ablation study. Specifically, for each network, we extract three layers of features $F_i, i=1,2,3$ from shallow layer to deep layer successively. The details of the features for each network are in the Supplementary. To eliminate the influence of the disparity loss $\ell_d$, we only use the warping loss $\ell_w$ to generate adversarial noise in this experiment. Table \ref{tab:warp123} compares the performance of the one-shot attack algorithm using $F_i, i=1,2,3$, respectively. For the target model CREStereo \cite{li2022practical} and DeepPruner \cite{duggal2019deeppruner}, the noise obtained by warping the shallowest feature $F_1$ results in the most significant error on the predicted disparity. For AANet \cite{xu2020aanet}, warping feature $F_1$ achieves the largest error on the metric D1-error, while warping feature $F_2$ achieves the largest error on the other two metrics. Such results probably indicate that warping a shallower feature can lead to more significant error on stereo networks. This is perhaps because during the forward propagation of a stereo network, the dissimilarity between left and right features of a shallow layer will propagate to deep layers. Since our warping loss $\ell_w$ tends to maximize the dissimilarity between the left and right features, when the warping loss is applied on a shallow feature, it will achieve the largest dissimilarity on the shallow feature, and such a large dissimilarity will propagate to the subsequent deep layers, causing the network to make large error on the predicted disparity.

\begin{table}[htbp]
    \centering
    \caption{Result comparison using three different layers of intermediate features $F_i, i=1,2,3$ (from shallow to deep) in the warping loss $\ell_w$. ``\textbf{A}" represents AANet, ``\textbf{D}" represents DeepPruner, ``\textbf{C}" represents CREStereo.}

    \setlength\tabcolsep{2pt} 
    \begin{tabular}{cccccccc}
    \multirow{2}{*}{\textbf{Target}}      & \multirow{2}{*}{\textbf{Feature}} & \multicolumn{3}{c}{\textbf{KITTI 2015}}                 & \multicolumn{3}{c}{\textbf{Scene Flow}}              \\
                                          &                                   & \textbf{MAE}  & \textbf{RMSE}  & \textbf{D1} & \textbf{MAE}   & \textbf{RMSE}  & \textbf{D1} \\
                                          \hline
    \multirow{3}{*}{\textbf{A}}       & \textbf{$F_1$}                       & 3.43          & 5.78           & \textbf{27.95}    & 9.00           & 13.73          & \textbf{54.97}    \\
                                          & \textbf{$F_2$}                       & \textbf{4.97} & \textbf{9.83}  & 26.70             & \textbf{9.52}  & \textbf{14.64} & 52.69             \\
                                          & \textbf{$F_3$}                       & 3.23          & 6.41           & 22.98             & 5.00           & 9.65           & 27.60             \\
                                          \hline
    \multirow{3}{*}{\textbf{D}} & \textbf{$F_1$}                       & \textbf{5.66} & \textbf{10.47} & \textbf{29.34}    & \textbf{15.48} & \textbf{26.77} & \textbf{67.77}    \\
                                          & \textbf{$F_2$}                       & 2.61          & 4.46           & 16.30             & 7.26           & 13.06          & 51.00             \\
                                          & \textbf{$F_3$}                       & 3.10          & 5.49           & 21.34             & 8.83           & 17.53          & 44.65             \\
                                          \hline
    \multirow{3}{*}{\textbf{C}}   & \textbf{$F_1$}                       & \textbf{1.68} & \textbf{2.57}  & \textbf{18.45}    & \textbf{2.27}  & \textbf{4.37}  & \textbf{20.05}    \\
                                          & \textbf{$F_2$}                       & 1.66          & 2.36           & 12.56             & 2.16           & 4.42           & 18.52             \\
                                          & \textbf{$F_3$}                       & 1.57          & 2.31           & 6.04              & 1.66           & 4.34           & 11.41             \\
    \end{tabular}

    \label{tab:warp123}
\end{table}

% \subsection{Transferability of our White-box Attack across Models}
% In order to investigate the transferability of the proposed method, the noisy image optimized by attacking one stereo network is feed into other stereo networks to generate error prediction results. 

\subsection{Effect of White-box Attack on Feature Map}
\label{subsec:fea}
\begin{figure}[htbp]
\centering

\subfigure[CREStereo]{\includegraphics[width=0.8\linewidth]{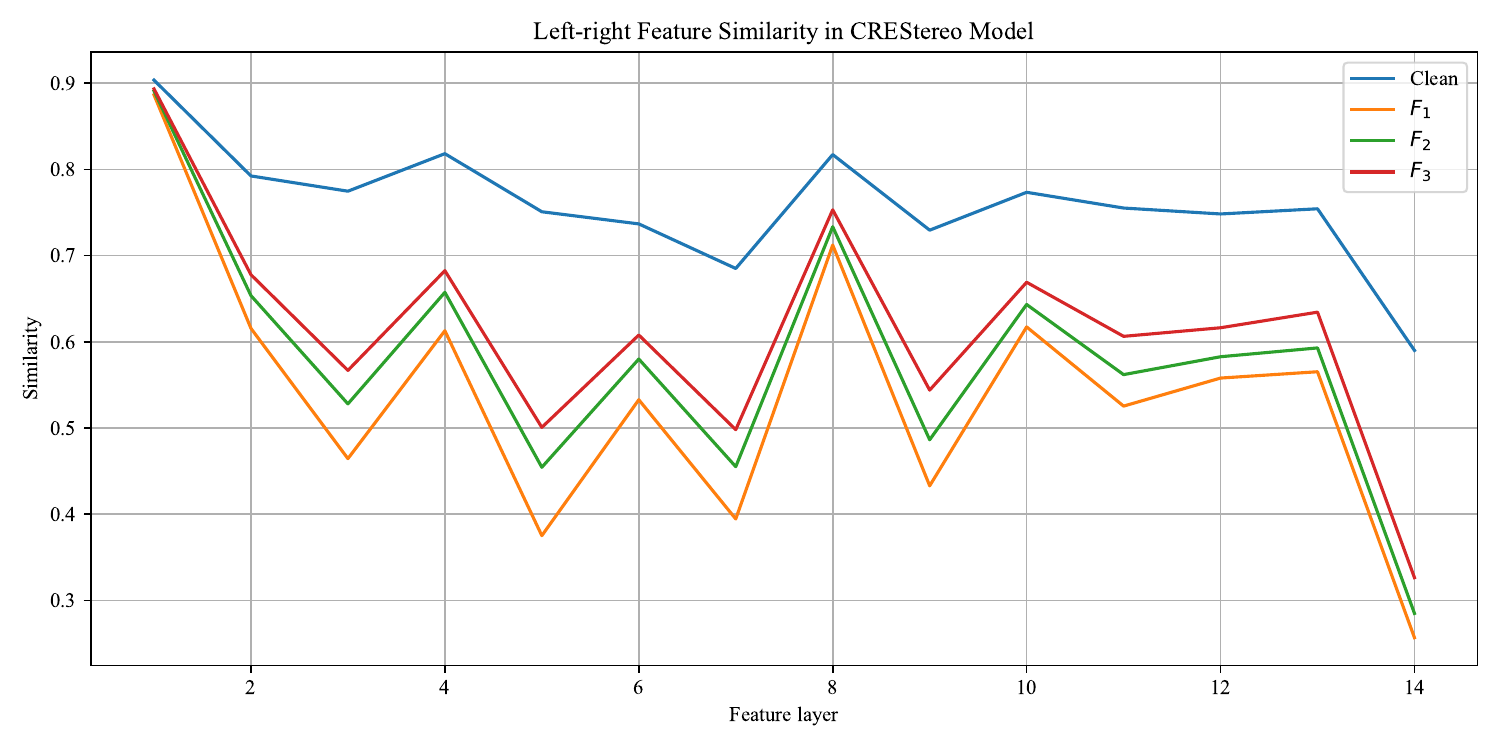}}\label{fig:1(a)}

\subfigure[AANet]{
\includegraphics[width=0.8\linewidth]{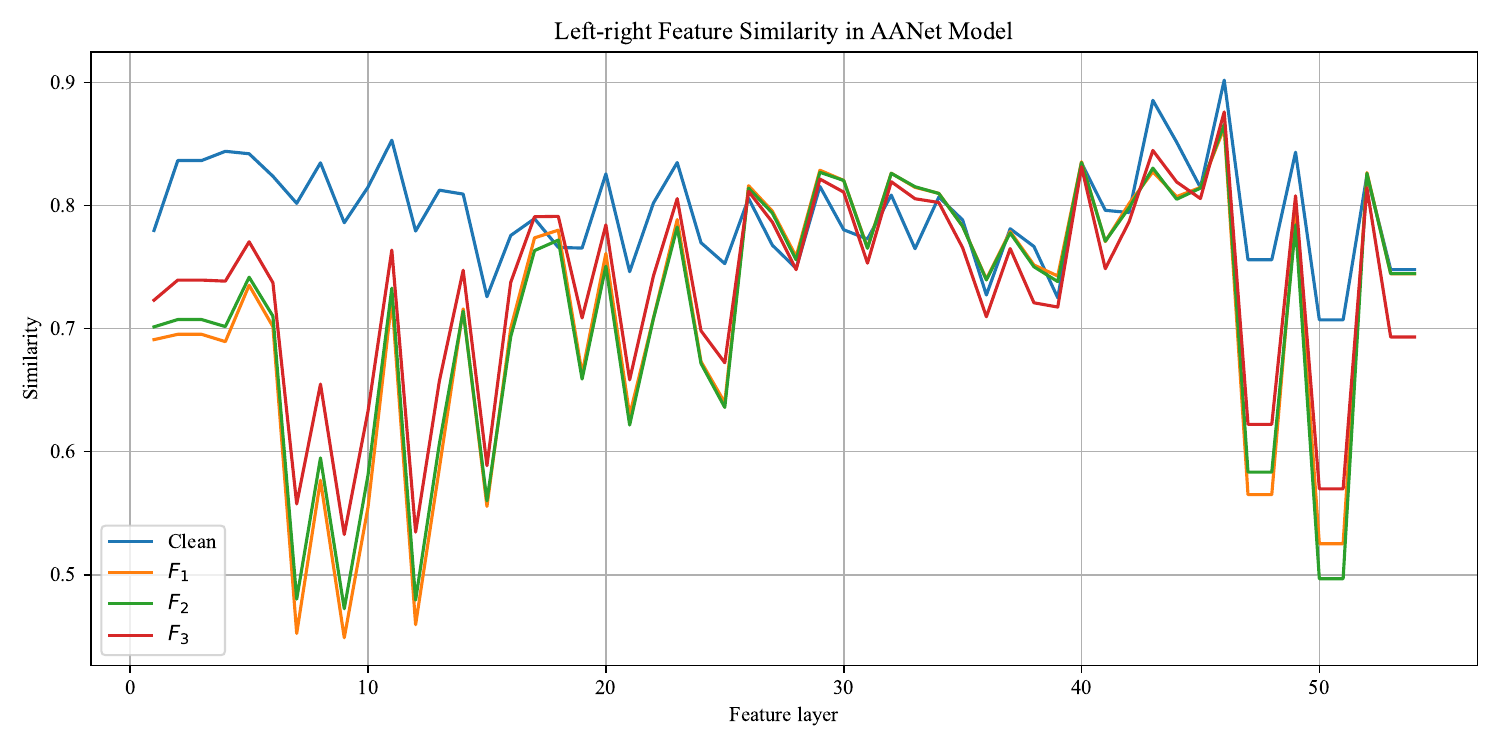}} 
\label{fig:1(b)}

\subfigure[DeepPruner]{\includegraphics[width=0.8\linewidth]{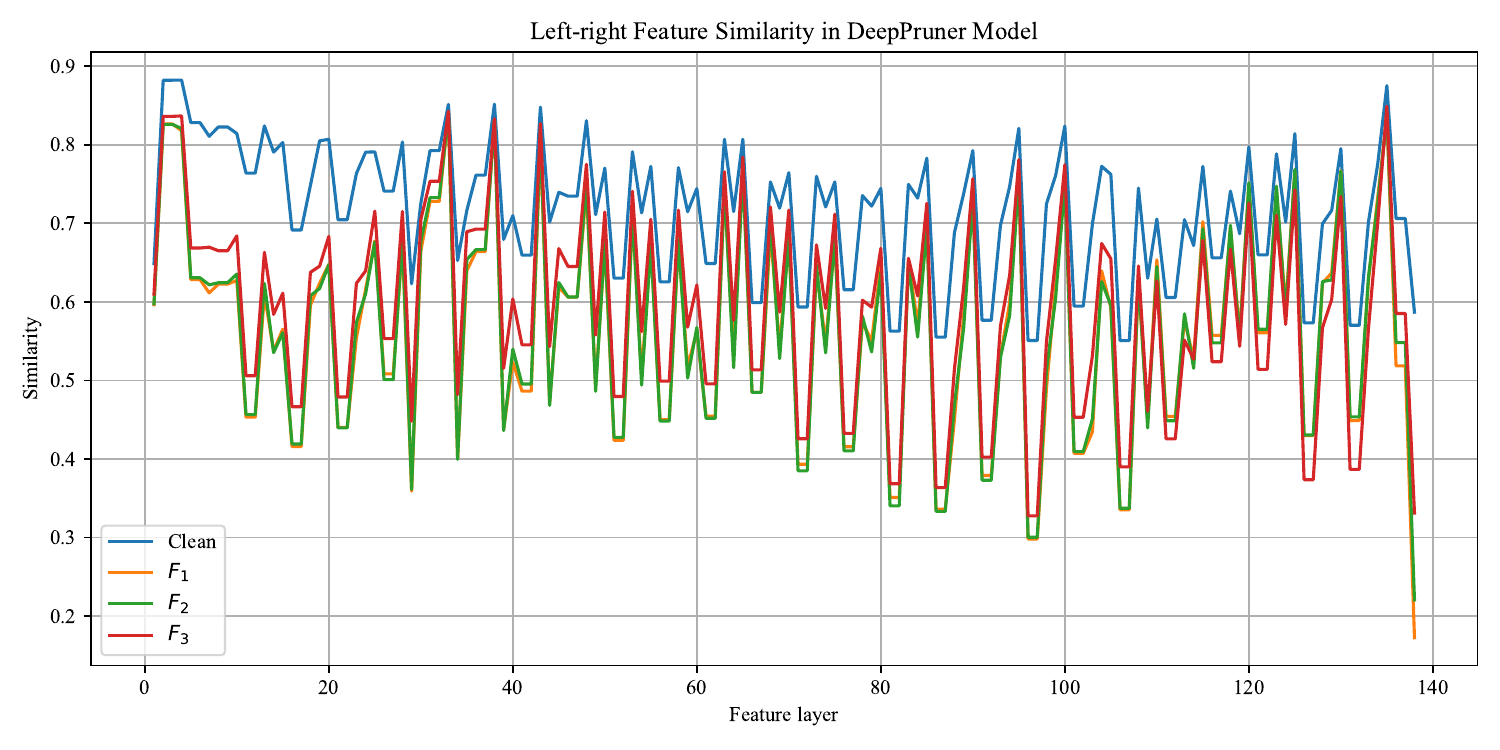}}
\label{fig:1(c)}

\caption{Left-right feature similarity before and after white-box adversarial attack. ``Clean" represents the similarity values without adversarial attack. $F_i, i=1,2,3$ represents the similarity values by applying warping loss using $F_i, i=1,2,3$ respectively.}
    \label{fig:sim_white}
\end{figure}

% As we assume the left-right feature similarities are crucial for the performance of stereo networks, and we propose generating adversarial noise to maximize the dissimilarity between the left-right features.
Empirical findings from prior research \cite{berger2022stereoscopic} has provided empirical evidence that adversarial attack on stereo networks can reduce the left-right similarity of intermediate features. Such observation aligns seamlessly with our assumption that the performance of the stereo network is significantly influenced by the left-right feature. To quantitatively analyze the impact of the proposed white-box attack method, we measure the correlation between the left and right feature similarities before and after adding adversarial noise. Particularly, in the feature layers with weight sharing between the left and right images, the intermediate feature following each convolution layer $f_{L,i}, f_{R,i}, i=1, 2, 3 ...$ is extracted. Subsequently, the right feature $f_{R,i}$ is warped to a pseudo left feature $f_{R_w,i}$. The similarity $S_i$ between the $i$-th left feature and pseudo left feature is calculated by the dot product $\frac{f_{L, i}}{||f_{L, i}||} \cdot \frac{f_{R_w, i}}{||f_{R_w, i}||}$. Figure \ref{fig:sim_white} shows the left-right similarity of each convolution feature for the three stereo networks. Obviously, the left-right similarity drops significantly for all the three stereo networks after adversarial attack. The average left-right similarity of all the layers is 0.715 for DeepPruner without adversarial attack, which drops to 0.563 (21.2\% drop rate) when applying the warping loss using $F_1$. Similarly, the average similarity value drop from 0.793 to 0.712 (10.2\% drop rate) for AANet, and from 0.759 to 0.539 (29.0\% drop rate) for CREStereo. Likewise, by warping feature $F_2$ or $F_3$, the results exhibit similar trend of left-right similarity drop. Such results demonstrate the effectiveness of our proposed warping loss. Through maximizing the left and right dissimilarity of an arbitrary intermediate feature, the similarity of the weight sharing left-right features across the whole stereo network drops significantly, leading to substantial prediction error. In addition, the drop rate of using $F_1$ in the warping loss is larger than that of $F_2$ and $F_3$. Such observation provides additional evidence for our assumption that by applying warping loss on a shallow feature, the large dissimilarity can propagate to deep layers, cause the stereo network to make a larger prediction error.

\begin{table*}[htbp]
    \centering
    \caption{Result of Black-box attack, $F'_i, i=1,2,3,4$ represents the intermediate features (from shallow layer to deep layer) used for the loss $\ell'$. Target stereo networks for adversarial attack are AANet, DeepPruner and CREStereo. The attack results are evaluated with three metrics. MAE-Inc, RMSE-Inc and D1-Inc indicate the percentage of increase compared to the results without adversarial attack.}

\begin{tabular}{clllllll}
\multirow{2}{*}{\textbf{Target}}      & \multicolumn{1}{c}{\multirow{2}{*}{\textbf{Feature}}} & \multicolumn{3}{c}{\textbf{KITTI 2015}}                                                                                                                                   & \multicolumn{3}{c}{\textbf{Scene Flow}}                                                                                                                                   \\
                                      & \multicolumn{1}{c}{}                                  & \multicolumn{1}{c}{\textbf{MAE-Inc\textbackslash\%}} & \multicolumn{1}{c}{\textbf{RMSE-Inc\textbackslash\%}} & \multicolumn{1}{c}{\textbf{D1-Inc\textbackslash\%}} & \multicolumn{1}{c}{\textbf{MAE-Inc\textbackslash\%}} & \multicolumn{1}{c}{\textbf{RMSE-Inc\textbackslash\%}} & \multicolumn{1}{c}{\textbf{D1-Inc\textbackslash\%}} \\
\hline
\multirow{4}{*}{\textbf{AANet}}       & $F'_1$                                                    & \textbf{121.15}                                         & \textbf{71.93}                                          & \textbf{506.66}                                       & \textbf{82.51}                                          & \textbf{36.33}                                          & \textbf{279.12}                                       \\
                                      & $F'_2$                                                    & 91.9                                                    & 58.28                                                   & 348.48                                                & 54.79                                                   & 28.85                                                   & 149.82                                                \\
                                      & $F'_3$                                                    & 74.3                                                    & 48.37                                                   & 268.7                                                 & 47.31                                                   & 26.39                                                   & 121.7                                                 \\
                                      & $F'_4$                                                    & 56.09                                                   & 34.84                                                   & 193.3                                                 & 44.13                                                   & 25.44                                                   & 114.18                                                \\
\hline
\multirow{4}{*}{\textbf{DeepPruner}} & $F'_1$                                                    & \textbf{127.95}                                         & \textbf{82.32}                                          & \textbf{504.14}                                       & \textbf{99.94}                                          & \textbf{58.2}                                           & \textbf{205.32}                                       \\
                                      & $F'_2$                                                    & 97.05                                                   & 65.07                                                   & 344.53                                                & 81.58                                                   & 51.3                                                    & 149.31                                                \\
                                      & $F'_3$                                                    & 74.76                                                   & 47.53                                                   & 251.06                                                & 72.52                                                   & 46.26                                                   & 132.31                                                \\
                                      & $F'_4$                                                    & 59.84                                                   & 38.75                                                   & 179.13                                                & 69.87                                                   & 45.07                                                   & 126.16                                                \\
\hline
\multirow{4}{*}{\textbf{CREStereo}}   & $F'_1$                                                    & \textbf{38.14}                                          & \textbf{12.6}                                           & \textbf{78.52}                                        & \textbf{39.42}                                          & \textbf{15.21}                                          & \textbf{111.91}                                       \\
                                      & $F'_2$                                                    & 11.36                                                   & 2.33                                                    & 18.67                                                 & 28.01                                                   & 15.05                                                   & 72.45                                                 \\
                                      & $F'_3$                                                    & 6.79                                                    & 0.45                                                    & 6.82                                                  & 24.79                                                   & 14.71                                                   & 60.07                                                 \\
                                      & $F'_4$                                                    & 7.28                                                    & 1.07                                                    & 9.44                                                  & 21.12                                                   & 12.59                                                   & 56.48                                                
\end{tabular}
\label{tab:black}
\end{table*}

\begin{table*}[htbp]
    \centering
    \caption{Result of Black-box attack by warping loss using random half feature channels, $F'_i, i=1,2,3,4$ represents the intermediate features (from shallow layer to deep layer) used for the loss $\ell'$.}

\begin{tabular}{clllllll}
\multirow{2}{*}{\textbf{Target}}      & \multicolumn{1}{c}{\multirow{2}{*}{\textbf{Feature}}} & \multicolumn{3}{c}{\textbf{KITTI 2015}}                                                                                                                                   & \multicolumn{3}{c}{\textbf{Scene Flow}}                                                                                                                                   \\
                                      & \multicolumn{1}{c}{}                                  & \multicolumn{1}{c}{\textbf{MAE-Inc\textbackslash\%}} & \multicolumn{1}{c}{\textbf{RMSE-Inc\textbackslash\%}} & \multicolumn{1}{c}{\textbf{D1-Inc\textbackslash\%}} & \multicolumn{1}{c}{\textbf{MAE-Inc\textbackslash\%}} & \multicolumn{1}{c}{\textbf{RMSE-Inc\textbackslash\%}} & \multicolumn{1}{c}{\textbf{D1-Inc\textbackslash\%}} \\
\hline
\multirow{4}{*}{\textbf{AANet}}       & $F'_1$                                                    & \textbf{136.14}                                         & \textbf{83.81}                                          & \textbf{589.67}                                       & \textbf{89.1}                                           & \textbf{39.74}                                          & \textbf{304.81}                                       \\
                                      & $F'_2$                                                    & 86.83                                                   & 55.51                                                   & 327.35                                                & 56.32                                                   & 30.26                                                   & 151.7                                                 \\
                                      & $F'_3$                                                    & 70.78                                                   & 44.76                                                   & 254.07                                                & 46.27                                                   & 25.98                                                   & 119.39                                                \\
                                      & $F'_4$                                                    & 55.21                                                   & 33.97                                                   & 183.53                                                & 44.23                                                   & 25.68                                                   & 114.18                                                \\
\hline
\multirow{4}{*}{\textbf{DeepPruner}} & $F'_1$                                                    & \textbf{146.84}                                         & \textbf{96.08}                                          & \textbf{592.96}                                       & \textbf{106.07}                                         & \textbf{63.01}                                          & \textbf{213.76}                                       \\
                                      & $F'_2$                                                    & 91.23                                                   & 60.4                                                    & 329.46                                                & 87.45                                                   & 56.44                                                   & 157.78                                                \\
                                      & $F'_3$                                                    & 71.94                                                   & 46.64                                                   & 248.3                                                 & 71.42                                                   & 45.71                                                   & 129.9                                                 \\
                                      & $F'_4$                                                    & 57.44                                                   & 36.6                                                    & 172.51                                                & 69.68                                                   & 44.84                                                   & 126.03                                                \\
\hline
\multirow{4}{*}{\textbf{CREStereo}}   & $F'_1$                                                    & \textbf{36.43}                                          & \textbf{12.35}                                          & \textbf{76.62}                                        & \textbf{39.65}                                          & \textbf{16.3}                                           & \textbf{118.63}                                       \\
                                      & $F'_2$                                                    & 12.54                                                   & 3.98                                                    & 20.64                                                 & 25.93                                                   & 14.16                                                   & 68.01                                                 \\
                                      & $F'_3$                                                    & 6.83                                                    & 0.62                                                    & 7.84                                                  & 22.79                                                   & 14.05                                                   & 60.72                                                 \\
                                      & $F'_4$                                                    & 6.79                                                    & 0.69                                                    & 8.45                                                  & 21.74                                                   & 13.1                                                    & 57.43                                                
\end{tabular}
\label{tab:black_random}
\end{table*}

\subsection{Performance of Black-box Attack}
\label{subsec:black}
As the target stereo network to be attacked is not always available, we propose a novel proxy network attack method by generating adversarial noise on a proxy network, afterwards, the noise is aggregated onto the input images to mislead the stereo network. In the experiment, we take the classical ResNet-18 classification network \cite{he2016deep} as the proxy to generate adversarial noise. Concretely, the left and right image are feed into the ResNet-18 network successively to extract intermediate features. Subsequently, FGSM \cite{goodfellow2014explaining} or I-FGSM is used to generate adversarial noise by maximizing the warping loss $\ell_w$, increasing the dissimilarity between the left and right features. In order to investigate how the adversarial attack performance is impacted by adopting different intermediate features in the warping loss, we extract four intermediate features ($F'_i, i=1,2,3,4$ from shallow layer to deep layer) from ResNet-18 in different scales for both left and right images. The details of the features are included in the Supplementary. As shown in Table \ref{tab:black}, CREStereo \cite{li2022practical} is more robust than AANet \cite{xu2020aanet} and DeepPruner \cite{duggal2019deeppruner} against the adversarial noise generated by our proxy black-box attack method. This trend of robustness is similar to that of white-box attack. Another noteworthy finding is that for all three stereo networks, the prediction error generated by warping different layers of ResNet features decreases monotonically from the shallowest layer to the deepest layer. Such results indicate that by maximizing the dissimilarity between the left and right features in a shallow layer of the proxy network, a more obvious error on the predicted disparity is achieved on the target stereo network. One possible explanation is that the shallow layers extract some low-level features in both the stereo disparity estimation or other tasks. Such low-level features share common knowledge across tasks, facilitating effective transfer of the dissimilarity from other tasks to the stereo network.

\begin{figure}[htbp]
\centering
\subfigure[CREStereo]{\includegraphics[width=0.8\linewidth]{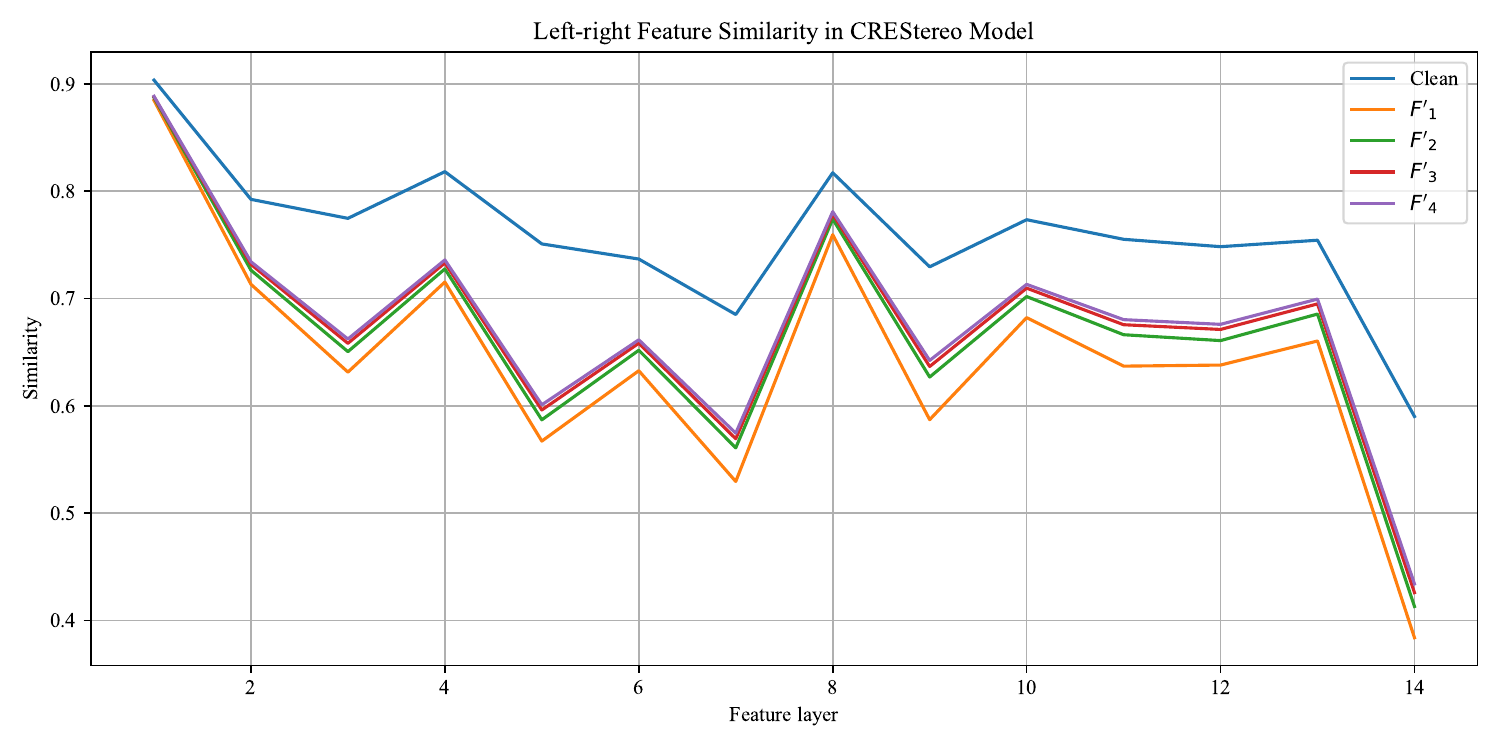}} \label{fig:1(a)}

\subfigure[AANet]{\includegraphics[width=0.8\linewidth]{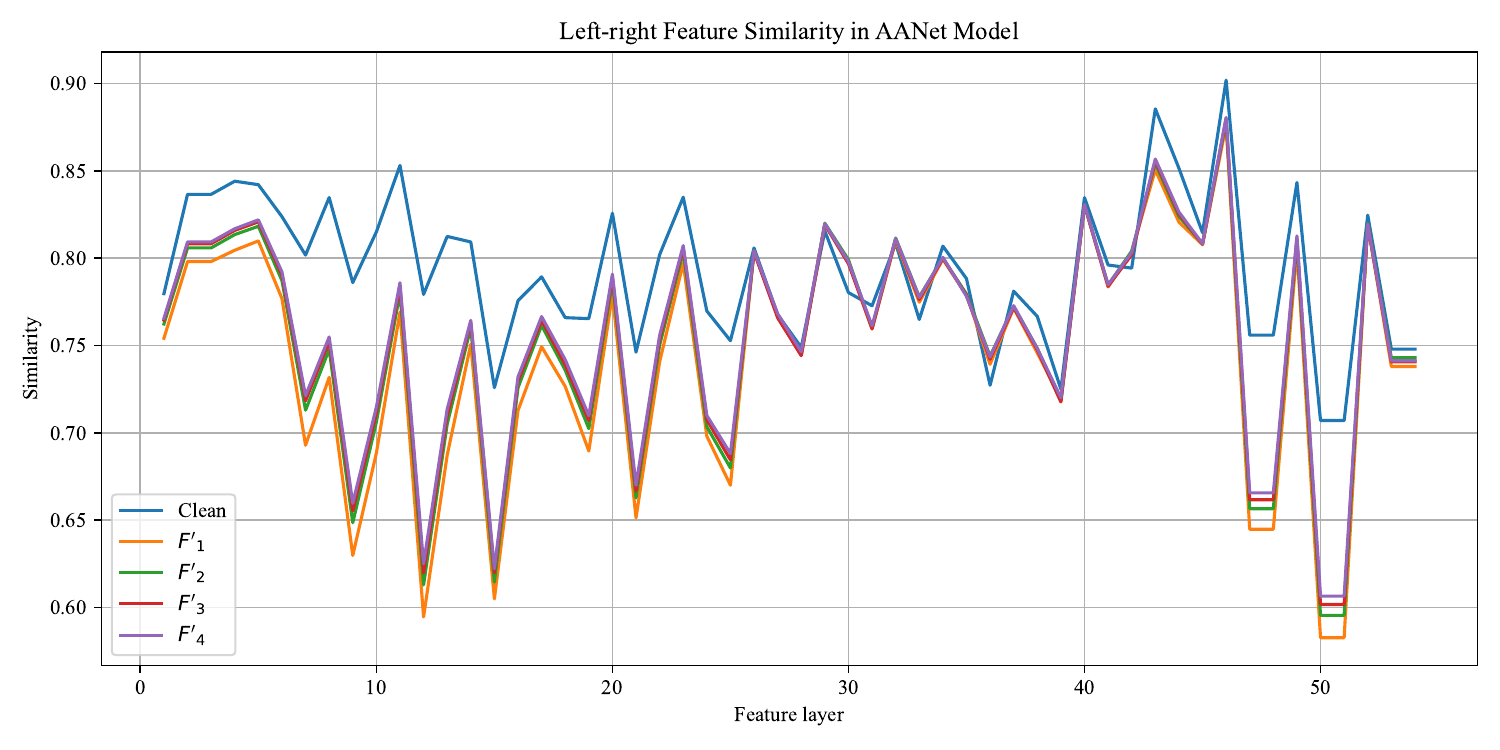}} \label{fig:1(b)}

\subfigure[DeepPruner]{\includegraphics[width=0.8\linewidth]{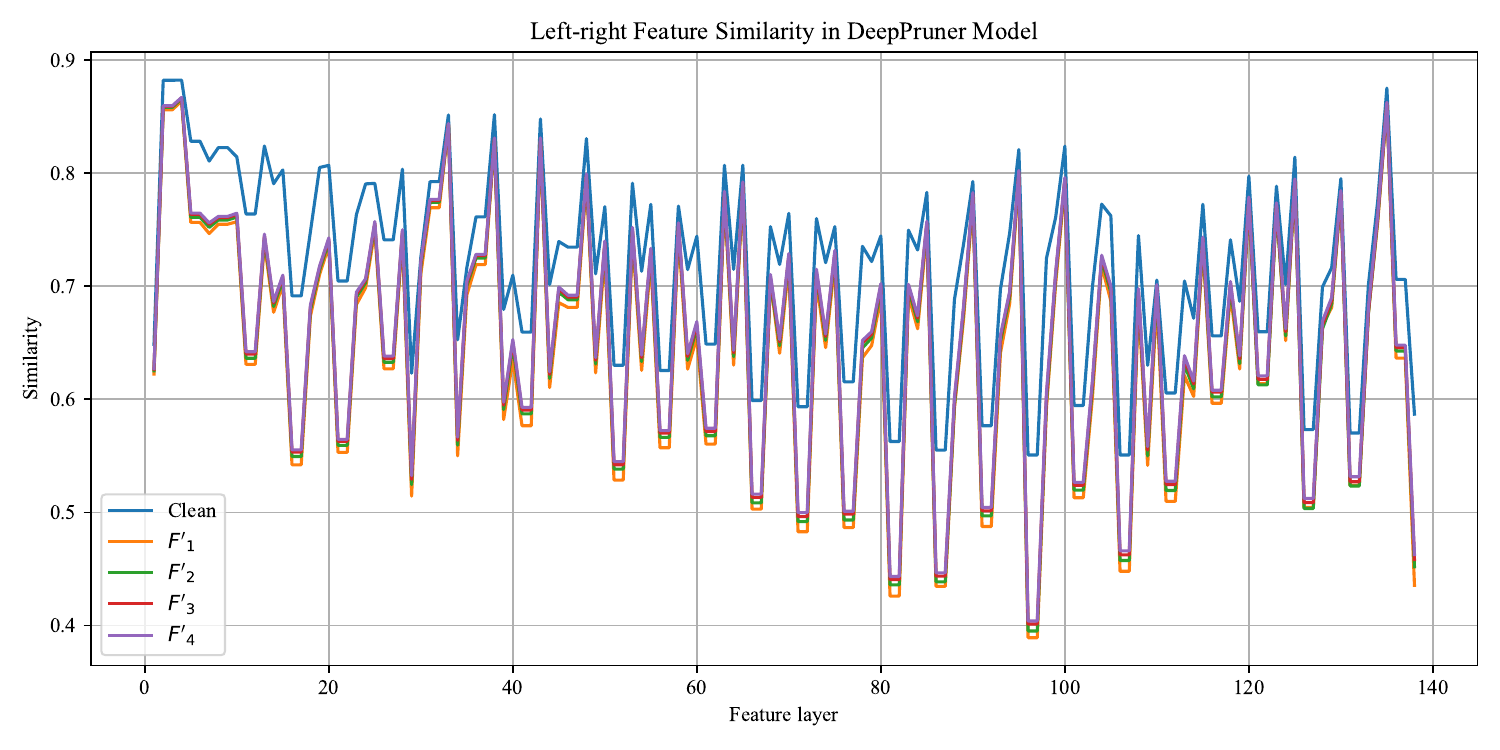}} \label{fig:1(c)}

\caption{Left-right feature similarity before and after black-box adversarial attack. ``Clean" represents the similarity values without adversarial attack. $F'_i, i=1,2,3, 4$ represents the similarity values by applying warping loss using the intermediate feature $F'_i, i=1,2,3, 4$ respectively.}
    \label{fig:sim_black}
\end{figure}

\begin{figure}[htbp]
\centering

\subfigure[CREStereo]{\includegraphics[width=0.8\linewidth]{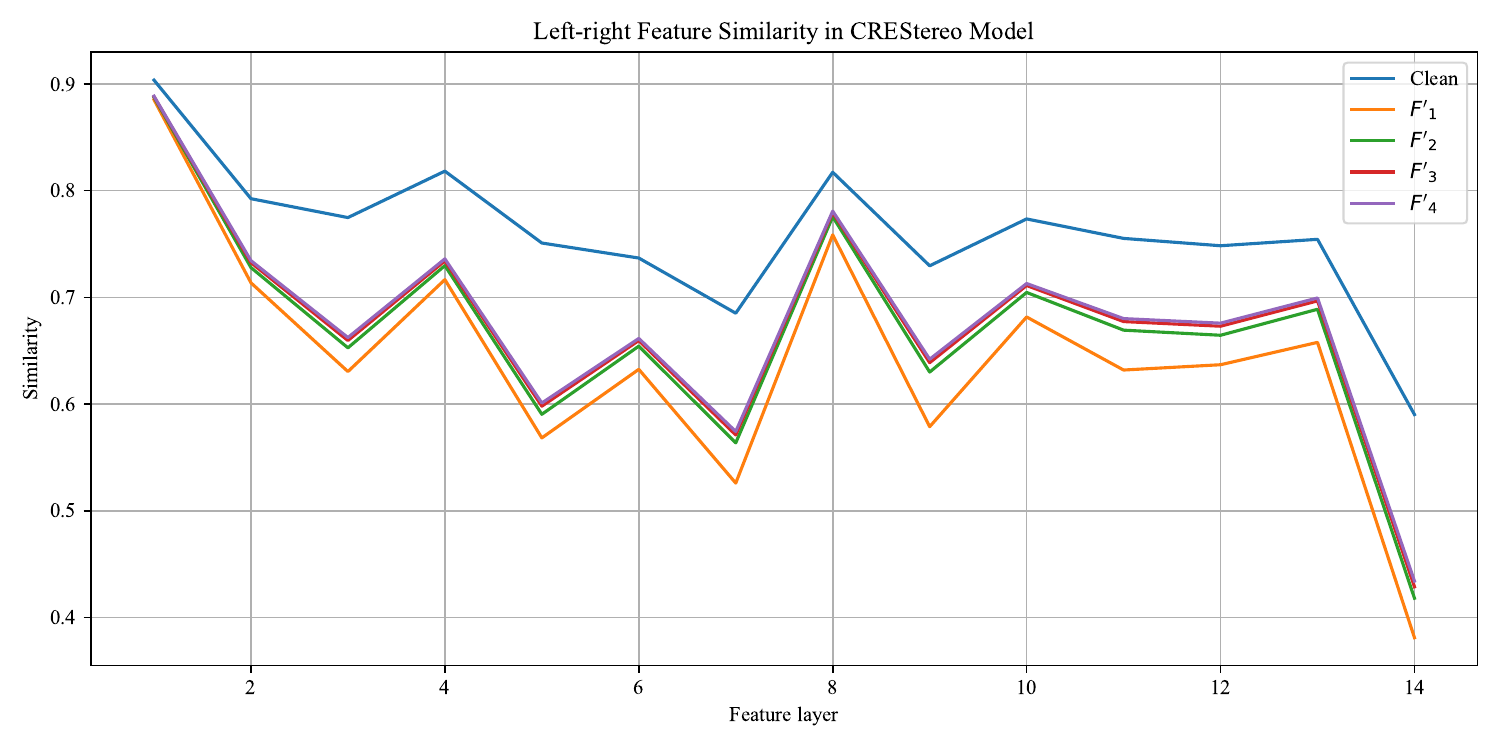}} \label{fig:1(a)}

\subfigure[AANet]{\includegraphics[width=0.8\linewidth]{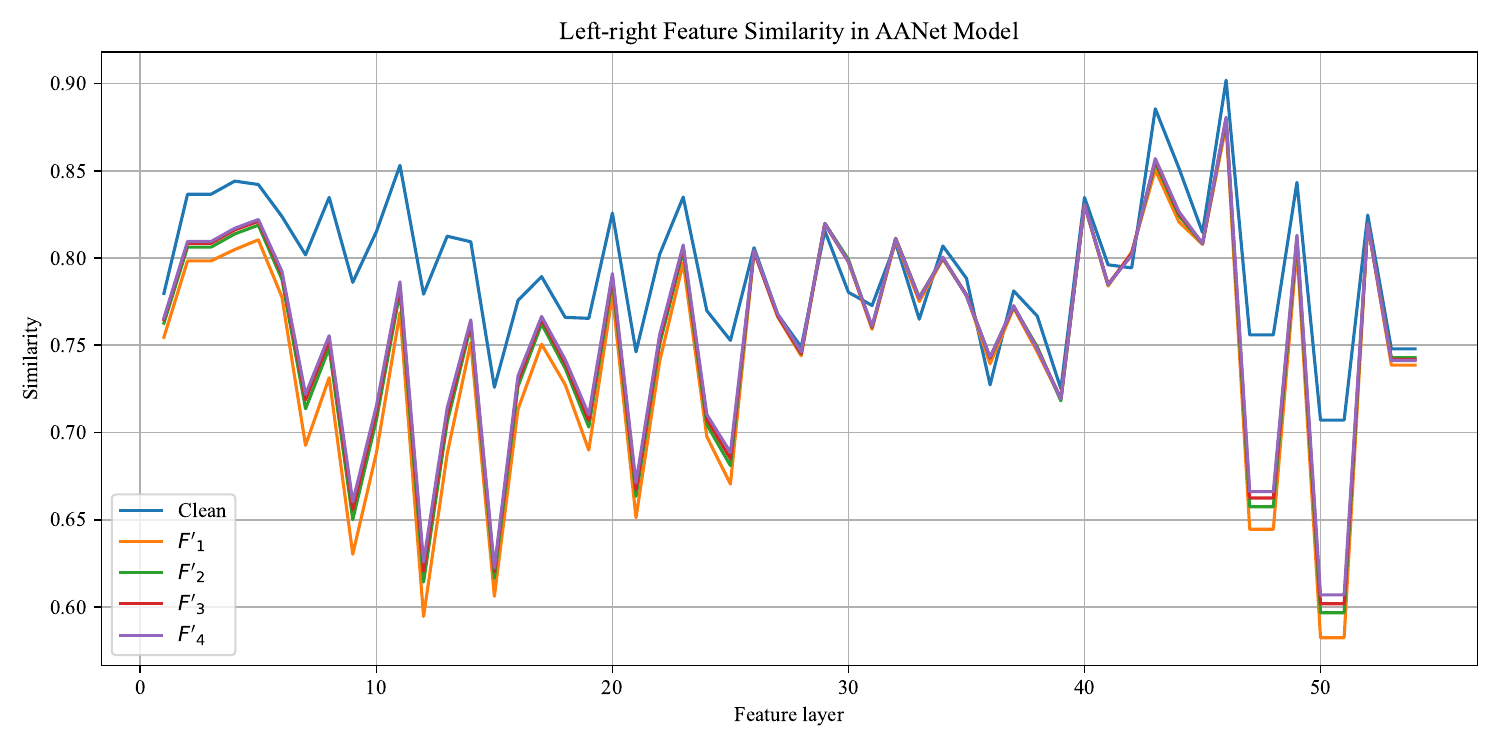}} \label{fig:1(b)}

\subfigure[DeepPruner]{\includegraphics[width=0.8\linewidth]{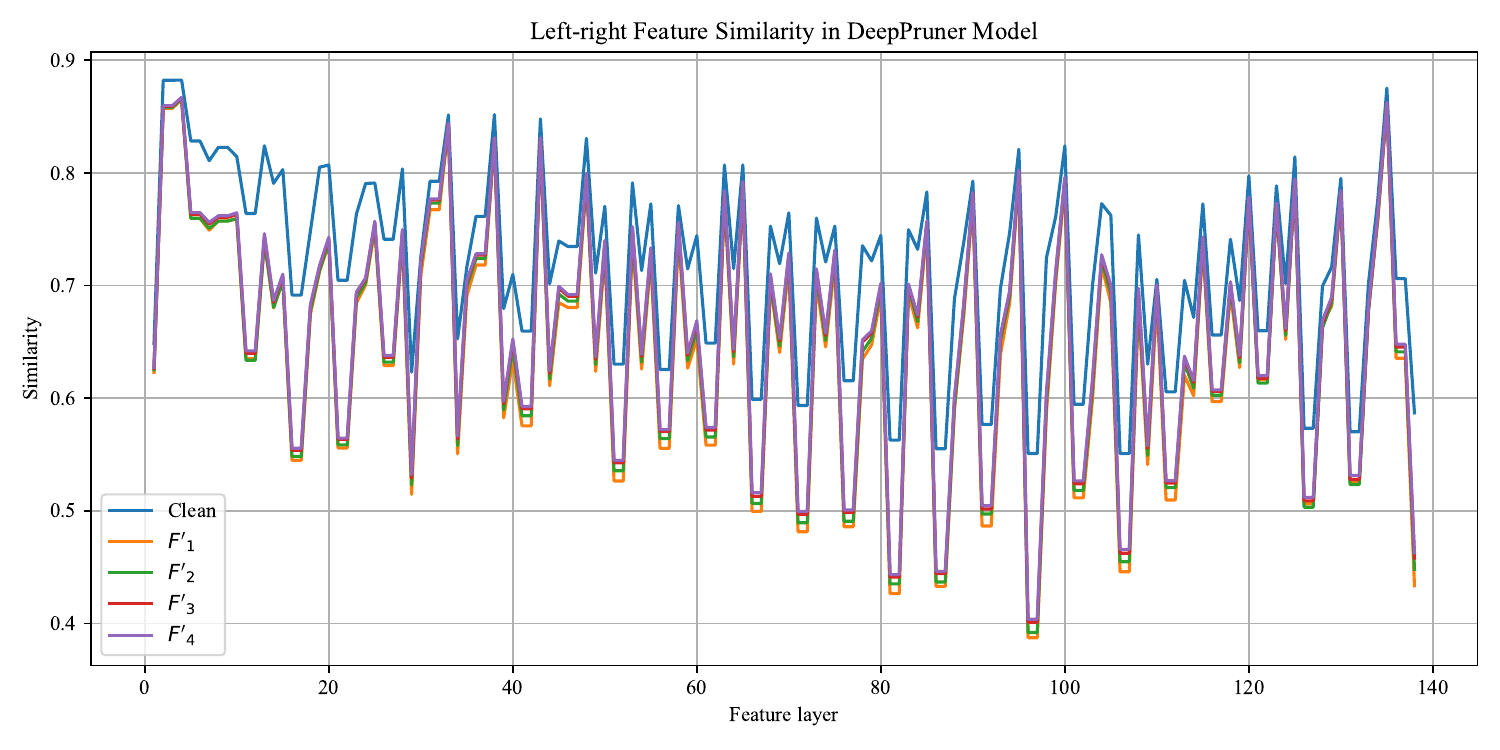}}\label{fig:1(c)}

\caption{Left-right feature similarity before and after black-box adversarial attack using random half feature channels. ``Clean" represents the similarity values without adversarial attack. $F'_i, i=1,2,3, 4$ represents the similarity values by applying warping loss using the intermediate feature $F'_i, i=1,2,3, 4$ respectively.}
    \label{fig:sim_black_random}
\end{figure}

% This shows the stereo networks are more sensitive to the noise on the shallow layers. Such results also indicate the 
To investigate in more detail the effect on intermediate features by warping different layer of features, the left-right similarity of each of the intermediate features is calculated. As shown in Figure \ref{fig:sim_black}, for all three stereo networks, warping the shallowest feature $F'_1$ causes the largest similarity drop. For example, for the CREStereo model, the average similarity drop after adversarial attack warping by warping $F'_i, i=1,2,3,4$ is 15.12\%, 12.33\%, 11.31\%, and 10.78\%, respectively. Such similarity drop caused by black-box attack shows a similar trend to that of white-box attack. By maximizing the dissimilarity of a shallow feature layer, the dissimilarity will propagate to the deep feature layers during the forward propagation of the stereo network, leading the network predict the most obvious error. This implies that the left-right similarity of shallow features is more critical than deep features to the robustness of stereo networks. 

% In our proxy model black-box attack method, the adversarial noise is generated through applying warping loss on an intermediate feature of ResNet. 
In order to further investigate the effectiveness of the warping loss, for the intermediate feature $F_i, i=1,2,3,4$, we randomly select half of the feature channels, and apply warping loss on the selected feature channels. An interesting finding is that by attacking the stereo networks using only half of the feature channels, it can achieve comparable error (see Table \ref{tab:black_random}) to that of using all the feature channels. This can reasonably be attributed to the convolution kernels within a convolution layer exhibiting a high degree of similarity and redundancy \cite{chen2019drop, chen2023run, han2020ghostnet}. The adversarial noise maximizing half of the feature channels can also cause significant similarity drop on other feature channels. This observation inspires us consider the possibility of simultaneous robustness enhancement and redundancy reduction for the stereo network.
Here, we also calculate the left-right similarity of the intermediate features, as shown in Figure \ref{fig:sim_black_random}. It also shows a similar trend of left-right similarity drop, which demonstrates the effectiveness of warping loss on attacking stereo networks.

\section{Conclusion}
\label{sec:conclusion}
In this paper, we have proposed a novel adversarial attack method on stereo networks. Considering that the similarity between the left and right features is crucial for stereo networks, our method generates noise by maximizing the discrepancy between the left and right intermediate features. Extensive experiments of adversarial attack on three stereo networks using two datasets have demonstrated the proposed method is more effective than the SOTA attack method. In addition, we also propose a proxy network method for black-box adversarial attack. Our proposed black-box method does not require the accessibility of any stereo network. We can use a plain backbone neural network, such as ResNet, as the proxy network and generate noise by maximizing the discrepancy between the intermediate features of the left and right images. Moreover, the experiments reveal that the stereo network tend to be more sensitive to the discrepancy of shallow layer features, which shedding light on future research of stereo network robustness improvement.

{\small
\bibliographystyle{ieee_fullname}
\bibliography{egbib}
}

\end{document}